%% file: main.tex
\documentclass[sigconf]{acmart}
\usepackage{amsmath}
\usepackage{amssymb}
\usepackage{geometry}
\usepackage{hyperref}
\usepackage{algorithm}
\usepackage{algorithmic}
\usepackage{enumitem}
\usepackage[flushleft]{threeparttable}

\begin{document}
\newcommand{\E}{\mathrm{E}}
\newcommand{\Expect}{{\rm I\kern-.3em E}}
\newcommand{\Var}{\mathrm{Var}}
\newcommand{\Cov}{\mathrm{Cov}}
\newcommand{\R}{{\rm\hbox{I\kern-.15em R}}}
\newcommand{\IR}{{\rm\hbox{I\kern-.15em R}}}
\newcommand{\IN}{{\rm\hbox{I\kern-.15em N}}}
\newcommand{\IZ}{{\sf\hbox{Z\kern-.40em Z}}}
\newcommand{\IS}{{\rm\hbox{S\kern-.45em S}}}
\newcommand{\Real}{I\!\!R}

\title{Improving Mild Cognitive Impairment Prediction via Reinforcement Learning and Dialogue Simulation}

\author{
Fengyi Tang$^1$, 
Kaixiang Lin$^1$, 
Ikechukwu Uchendu$^1$,
Hiroko H. Dodge$^{2,3}$, 
Jiayu Zhou$^1$}
\affiliation{%
  \institution{$^1$Computer Science and Engineering, Michigan State University\\
               $^2$Michigan Alzheimer\textquotesingle s Disease Center, Department of Neurology, University of Michigan\\
               $^3$Layton Aging and Alzheimer\textquotesingle s Disease Center, Department of Neurology, Oregon Health \& Science University\\
               }
}
\email{{tangfeng,linkaixi, uchendui}@msu.edu, hdodge@med.umich.edu, jiayuz@msu.edu}




\begin{abstract}
\input{abs.tex}

\end{abstract}



\renewcommand{\shortauthors}{F. Tang et al.}
\renewcommand{\shorttitle}{MCI Prediction via Reinforcement Learning and Dialogue Simulation}
\newcommand{\commlkx}[1]{{\color{blue}(Kaixiang's comments: #1)}} 

\newcommand{\vs}{{\mathbf{s}}}
\newcommand{\vx}{{\mathbf{x}}}
\newcommand{\va}{{\mathbf{a}}}
\newcommand{\vo}{{\mathbf{o}}}
\newcommand{\vw}{{\mathbf{W}}}
\newcommand{\vq}{{\mathbf{q}}}
\newcommand{\vv}{{\mathbf{v}}}
\newcommand{\cE}{{\mathcal{E}}}
\newcommand{\cR}{{\mathcal{R}}}
\newcommand{\RR}{{\mathbb{R}}}
\newcommand{\EE}{{\mathbb{E}}}
\newcommand{\eq}[1]{{Eq~(#1)}}

\maketitle

\section{Introduction}
\input{sec_intro.tex}

\section{Related Work}
\input{sec_related.tex}

\section{Methodology}
\input{sec_method.tex}

\section{Experiments}
\input{sec_exp.tex}

\section{Discussion and Conclusion}
In this paper, we introduce a RL framework for approaching a classically supervised learning problem in clinical medicine, where the data is often noisy, scarce, and prohibitively expensive to obtain. We show that a properly trained RL framework can (1) greatly cut down on the amount of data needed to make accurate predictions, and  (2) synthesize relevant new data to improve performance. 

To achieve this framework, we proposed a multi-step approach which capitalizes on the vast existing knowledge of the human language and NLP research. First, we used a state-of-art distributed representation to preprocess our data. 
We then set up a simulation environment for reinforcement learning using supervised learning to create customized user simulators. 
Lastly, we utilize the trained RL-agent to generate new questions from $\pi^*$ to obtain more targeted responses for our prediction task. 

A careful examination of the optimal policies discovered by our agent demonstrates that the overall framework is self-contained for directing dialogue generation for diagnostic screening, which can potentially replace the need for trained interviewers. Our trained RL-agent is able to discover relevant questions to ask users where the agent has no prior experience of interaction. We also show various clinical insights which could be deduced from observing the ranking of questions in $\pi^*$ at various turn constraints. 

In order for this framework to be effectively deployed in a realistic setting, a user-simulator that could be trained online and in real-time should be considered. In its current form, our user-simulators are trained offline, which may not be scalable to larger corpus and user volumes. Additionally, a natural language generator phase may be needed to make the questions more adaptable to the natural flow of human conversation. 
These will be areas of research we will explore in future studies. 

\begin{acks}
This material is based in part upon work supported by the 
National Science Foundation (IIS-1565596, IIS-1615597), 
National Institute on Aging (R01AG056102, R01AG051628, R01AG033581, P30AG008017, P30AG053760) and
Office of Naval Research (N00014-17-1-2265).

\end{acks}

\bibliographystyle{ACM-Reference-Format}
\bibliography{ref.bib}

\end{document}

%% file: abs.tex



Mild cognitive impairment (MCI) is a prodromal phase in the progression from
normal aging to dementia, especially Alzheimer's disease (AD). Even though
there are mild cognitive declines in MCI patients, they have normal overall
cognition and thus is challenging to distinguish from normal aging. 
Using transcribed data obtained from recorded conversational interactions between participants and trained interviewers, and applying supervised learning models to these data, a recent clinical trial has shown a promising result in differentiating MCI from normal aging.
However, the substantial amount of interactions with medical staff can still incur
significant medical care expenses in practice. In this paper, we propose a novel
reinforcement learning (RL) framework to train an efficient dialogue agent on
existing transcripts from clinical trials. Specifically, the agent is trained
to sketch disease-specific lexical probability distribution, and thus to converse in a
way that maximizes the diagnosis accuracy and minimizes the number of
conversation turns. We evaluate the performance of the proposed RL framework
on the MCI diagnosis from a real clinical trial. The results show that while using only a few turns of conversation, 
our framework can significantly outperform state-of-the-art supervised learning
approaches.

%% file: sec_intro.tex

The progression of Alzheimer Disease (AD) has consistently been a heavy area of research in clinical medicine 
because while the disease itself is incurable, early intervention at the prodromal phases of the disease has proven to delay the onset of AD-related mental degeneration and systemic issues for months to years~\cite{olazaran2004benefits,cummings2007disease}.
Consequently, much of the recent
clinical research efforts have focused on detecting early stages of mild
cognitive impairment (MCI), which is a prodromal phase in AD progression
occurring months to years before visible mental decline
begins~\cite{gauthier2006mild}. If successfully detected at this stage,
intervention methods may confer numerous benefits in the longevity of
cognitive and physiological health of AD patients~\cite{olazaran2004benefits,cummings2007disease}.

Brain imaging, such as the structural magnetic resonance imaging (MRI), was
shown to contain prime markers of AD, capturing the physiologic changes in the
AD pathological process~\cite{johnson2012brain,heister2011predicting}.
However, the identification of MCI from normal aging (NL) is particularly
challenging due to the fact that structural changes in the brain in this phase
are minor and hard to detect through structural
MRI~\cite{jack2010hypothetical}, even though decline in mental
status and cognitive have already begun in most cases. Recently, the
structural connections among brain regions inferred from diffusion MRI have
provided promising predictive performance of MCI
detection~\cite{zhan2015boosting,wang2016discriminative}, yet sketching brain
networks via imaging still remains rather prohibitively expensive and
difficult to scale. Moreover, the high dimensionality of brain imaging
combined with small sample size usually imposes significant challenges in
learning algorithms and leads to unstable
generalization performance.

On the other hand, behavior and social markers could offer a much more cost-
effective option for MCI detection~\cite{dillon2013behavioral,chapman2011predicting,h2015social,asgari2017predicting}. A recent
clinical trial has studied differentiating early stage MCI from NL cohort
groups using transcripts of extensive conversations between patients and
trained interviewers~\cite{h2015social}. In a recent preliminary
study~\cite{asgari2017predicting}, the authors trained supervised learning
models from the \emph{lexical distribution} of the conversation, and showed that
conversational responses of MCI and NL patients take on different distribution
under various conversational topics. The success of
~\cite{asgari2017predicting} in predicting MCI using human dialogue introduced
an alternative natural language processing (NLP) approach to a classically
clinically expensive problem. However, the use of human interviewers still
requires substantial amounts of interaction between trained staff
which incur significant expense in its current form. Thus, the bottleneck
questions remain: \textit{(1) can we cut down on the amount of conversations
needed to achieve accurate prediction}, \textit{(2) can we improve upon
baseline performance given limited cohort-specific data}?

To address the aforementioned questions above, in this paper we propose a
novel reinforcement learning (RL) framework, which learns a MCI diagnosis
agent using only very limited amount of offline human dialogue transcripts.
The learned diagnosis agent can conduct subject-specific conversation with
humans, asking questions based on existing conversations to efficiently sketch
the lexical distribution and give high-performance MCI prediction. 
In order to facilitate RL using offline transcripts, we introduce a
\textit{dialogue simulator pipeline} which generates new conversational
episodes that are less noisy and out-perform the original corpus for MCI
prediction. 
Our dialogue pipeline provides a self-contained framework for directing dialogue generation for diagnostic screening which can potentially replace the need for human-expert interviews. 
Our RL-agent learns optimal dialogue strategies that are adaptive to unseen users, enabling medically-relevant NLP data to be generated on a large scale if deployed in a realistic setting.
Furthermore, data generated from our dialogue simulations may be
used for data augmentation or to perhaps guide the medical data collection process in
the future. Ultimately, by greatly decreasing the cost of data collection and the amount needed for high-level performance, we introduce a clinical
direction that is much more cost-effective and scalable to large-scale
diagnostic screening and data collection. The combination of
NLP features with our reinforcement learning framework
may extend the process of diagnostic screenings to well beyond the confines of
hospitals and primary care facilities.

%% file: sec_related.tex

\noindent\textbf{MCI Prediction via Utterance Data.}
\cite{asgari2017predicting} used a classical supervised learning framework to formulate MCI prediction as binary classification problem. For each interview, a corpus was constructed using only the participant responses to interviewer questions.
For each participant, the response corpus over several interviews was preprocessed into feature vectors using the Linguistic Inquiry $\&$ Word Count (LIWC) dictionary ~\cite{pennebaker2001linguistic}. 
The LIWC dictionary transforms each word in a given corpus to a 69-dimensional feature vector with latent dimensions representing grammatical and semantic properties of each word. 
A final 69-dimensional feature vector is then constructed at the end of the corpus by aggregation of all previous LIWC vectors. The resulting feature representation is a $m \times 69$ matrix. 
The best performing classifier in this benchmark study uses linear support vector machines (SVM) with $\ell_1$-norm regularization ~\cite{asgari2017predicting}. The resulting performance is 72.5\% AUC over 5-fold validation.

\noindent\textbf{Dialogue Systems.}
Dialogue systems provide a natural human-computer interface and have been an
active research field for decades. Task-oriented dialogue systems are
typically designed for retrieval- tasks in which users provide queries and the
chat-bot provides appropriate responses based on an external knowledge
base~\cite{wen2016network,dhingra2016end,chen2017survey}, or identifying
correct answers by looking into vast amounts of
documents~\cite{hu2017reinforced,he2017dureader}.
Such dialogue systems are typically
designed to be a pipeline containing a set of components including a language
understanding unit that parses the intention and semantics from the input from
humans, a dialogue manager that manages dialogue state tracking and policy
learning, and a language generation unit that generates
response~\cite{chen2017survey,schatzmann2006survey,singh2002optimizing}.
While each of the components can be handcrafted or trained individually,
recent advances of deep learning allows end-to-end
training~\cite{wen2016network,dhingra2016end,li2017end,li2017end} and
significantly improves the performance and the capability to adapt to
new domains~\cite{bordes2016learning}. The end-to-end systems can be 
trained using supervised learning~\cite{wen2016network,liu2017end} or 
reinforcement learning (RL), by leveraging a user simulator~\cite{dhingra2016end,li2017end}. 
The main advantage of RL is that less training samples are needed to 
learn the high-degree-of-freedom deep models. 
In our work, we design a simulator to enable RL due to the limited amount of clinical data available for supervised training. 
We note that even though our dialogue system also tries to achieve a task (identifying MCI
patients), the nature of our system is radically different from existing 
task-oriented dialogue systems: its goal is to efficiently sketch 
a disease specific lexical distribution through asking subject-specific questions and 
give classification results.

 

\noindent\textbf{Healthcare Applications of Dialogue Systems.}
Dialogue systems have been widely adopted in the healthcare domain for various applications. For example, chat-bots are available to assist the patient
intake process~\cite{ni2017mandy}, retrieve restaurant accommodation
information for young adults with food allergies~\cite{hsu2017allergybot}, and
perform dialogue analysis and generation conversation to perform mental health
treatment~\cite{oh2017chatbot}. In the context of Alzheimer's disease research,
\cite{montenegro2017cognitive} designed a virtual reality based chat-bot to
evaluate memory loss using predefined questions and answers.
\cite{salichs2016study} discussed applications of chat-bots as caregiviers for Alzheimer's patients, providing safety, personal assistance, entertainment,
and stimulation. More recently, \cite{tanaka2017detecting} introduced a
computer avatar to ask a list of pre-defined questions from neuropsychological
tests to diagnose dementia. This work is closely related to our system as it utilizes dialogue to glean disease-related information. However, one major issue in this approach is that the questions were obtained from mini-mental state examination (MMSE)~\cite{tombaugh1992mini}, which is
a confirmatory measure used to \emph{define} clinical dementia (such as MCI) rather than a diagnostic tool to predict it. 
It is more clinical meaningful to identify diagnostic markers associated with the pathological pathways, such as lexical distribution associated with the cognitive changes for the purpose of diagnostic screening.



%% file: sec_method.tex

\begin{figure}[t!]
\centering
\includegraphics[width=0.48\textwidth]{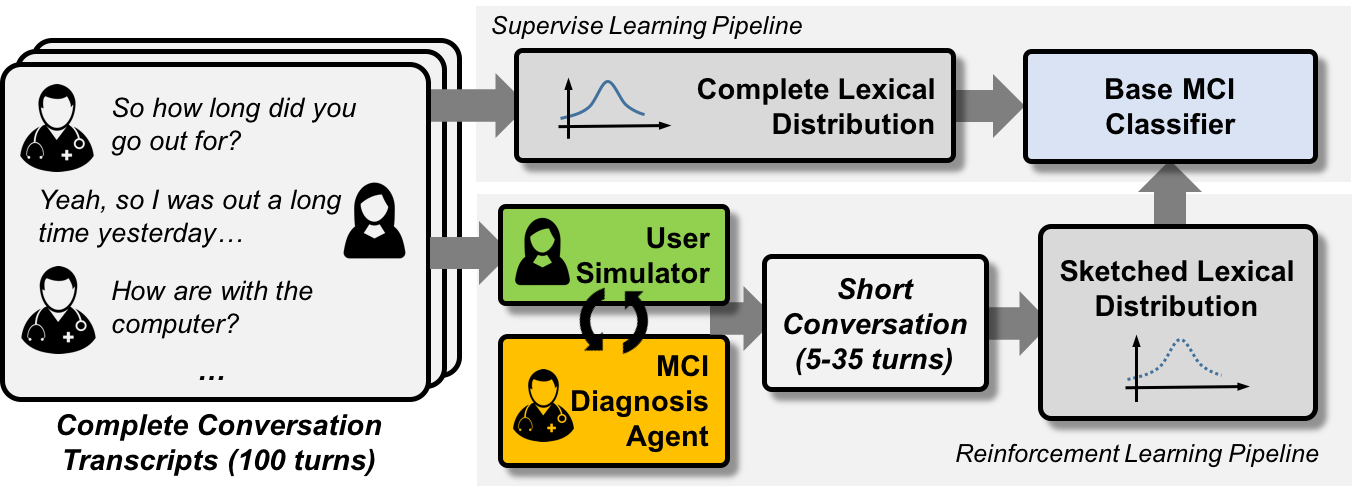}
\vspace{-0.25in}
\caption{Overview of the proposed methodology.
Complete conversation from participants are used to build user
simulators. The simulators are then used to train an MCI diagnosis 
agent (chat-bot), which conducts minimal turns of conversation with participants 
to sketches the lexical distribution that
is then used to perform MCI classification. }
\label{fig:fig3}
\vspace{-0.2in}
\end{figure}

The framework we propose in this paper involves the use of reinforcement learning to learn the optimal set of questions $\pi^*$ to ask participants for the purposes of distinguishing MCI. On test set, we generate new episodes from these questions for prediction rather than the original corpus. To actualize the RL + dialogue simulation framework, we proposed a multi-step approach for implementation which capitalizes on the vast existing knowledge of NLP research. In the following section, we present the details of each component of the dialogue system. 
Figure \ref{fig:fig3} shows an overview of the components of our experimental pipeline. 

\subsection{Overview of Pipeline}
\input{sec_pipline}

\begin{figure*}[t!]
\centering
\includegraphics[width=0.85\textwidth]{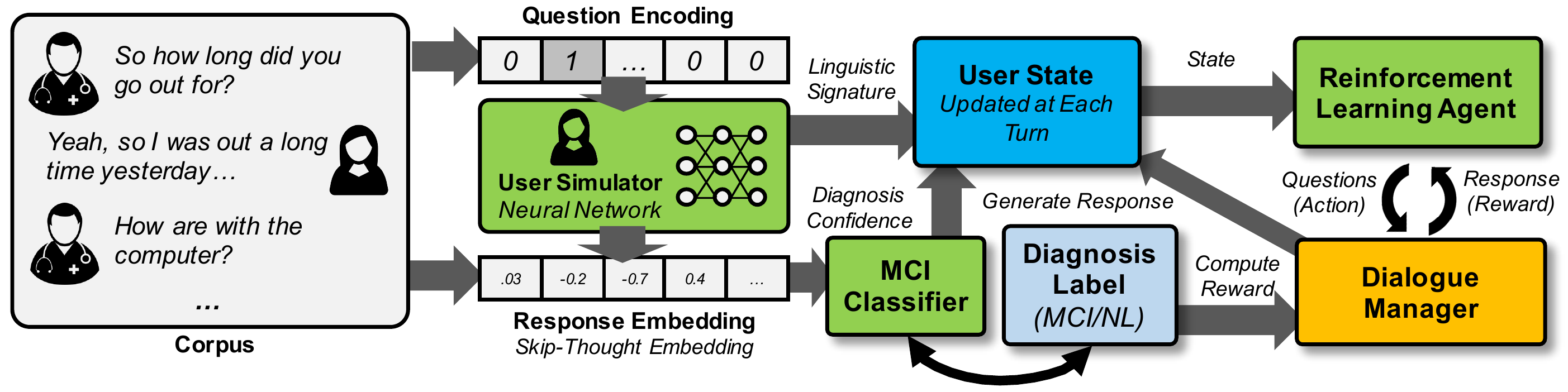}
\vspace{-0.15in}
\caption{Illustration of reinforcement learning components in our proposed approach.}
\label{fig:fig4}
\vspace{-0.15in}
\end{figure*}

\subsection{Construction of Turn-Based Dialogue}
Since utterance data was collected in the form of conversational transcripts for each participant, we must reconstructed turn-based dialogue from participant-responses. 
The participant responses were unstructured while interviewer questions ranged over preset question topics, as illustrated below.
\\ \\
\texttt{
\textbf{Interviewer}: so what did you do yesterday? \\
\textbf{Participant}: i had yesterday morning i yesterday was a busy day for me. i im forgetting i went to where did i go in the morning.
 well i went to albertsons yesterday...\\
\textbf{Interviewer}: what do you see in this picture?\\
\textbf{Participant}: we got a picture gosh. it looks like my uncle lou.
 but he never ...\\
\textbf{Interviewer}: when do you think this picture was taken?
\textbf{Participant}: this picture was probably eighteen seventy or something or nineteen twenty. so he looks too old for war he must have been ...}
\\ \\
In total there were well over 150 possible queries from the interviewers. However, for the purposes of this study, we re-compiled the question list into 107 general questions which were ubiquitous across all conversations. A snapshot of questions are in Table~\ref{tb:question-example}. 

\begin{table}[tb!]
\caption{Examples of questions from conversations:}
\label{tb:question-example}
\vspace{-0.1in}\small
\begin{tabular}{c|c}
\hline
\textbf{Category} & \textbf{Question}\\
\hline\hline
Activity &Did you go outside lately?\\
{}&So what did you do yesterday?\\
\hline
Social&Did you run into any familiar faces lately?\\
{}&Where did you have dinner?\\
\hline
Picture&What do you see in this picture?\\
{}&Where do you think this picture was taken?\\
\hline
Tech&How are you with the computer?\\
{}&Did you use your computer lately?\\
\hline
Unspecified&<unspecified scheduling comment>\\
{}&<unspecified picture comment>\\ 
\hline
\end{tabular}
\vspace{-0.2in}
\end{table}

We created a total of 16 question categories, including: \textit{greetings, activity check, living situation, travel, entertainment, social, picture-related, tech, occupation, hobbies, family, pets, confirmation, clarification, goodbye} and \textit{unspecified} comments. For some of these comments, we delexicalised certain topic words such as ``<\textit{activity}>'', ``<\textit{social topic}>'' in order to (1) control for domain expansion~\cite{henderson2014robust} and (2) reduce model complexity of our user simulators. In the past, ~\cite{henderson2014robust} and ~\cite{liu2017end} have shown the effectiveness of delexicalisation in controlling for domain expansion in user simulators without sacrificing the contextual meaning of sentence queries. Additionally, we also created \textit{unspecified comments} category, which included comments that deviated from general question prompts. These comments often result from interviewer follow-up on specific topics mentioned by the user. We consolidated these comments into a single category to distinguish the context-specific from general questions based on the corpus. However, we do demarcate the type of \textit{unspecified comment} used by the interviewer. For example, a follow-up comment to an occupational story is tagged \textit{<unspecified occupational comment>} whereas a follow-up comment about a health concern is tagged \textit{<unspecified health comment>}. The role of these comments serve to build rapport and improve flow of conversation. In future studies we may look to generate user-specific grounding statements for these slots~\cite{chai2016collaborative}. Implemented in this way, the corpus is tokenized into turn-based responses to questions for each user. 
\subsection{Unsupervised Learning for User Simulator}
To effectively capture contextual representation of user conversation style, 
we utilize vector embedding of user corpus at the sentence-level representation ~\cite{kiros2015skip,mikolov2013distributed}. 
Given that we want to capture the flow of the conversation from one response to the next,
we implement skip-thought embedding, which has shown effectiveness over large corporal datasets by capturing contextual information of sentences given neighboring ones ~\cite{kiros2015skip}. 
For encoding sentences, we use a model that was pretrained on the BookCorpus dataset, which contains turn-based conversations from various English novels ~\cite{kiros2015skip}. 
For the decoder, we train skip-thought vectors to recover the original response of the user during NLG portion of the pipeline.

Since each user has individual response styles to questions,  we train a
personalized user-simulator for each user. For each user,  the conversation
corpus is divided into question-response turns.  In our dataset, for example,
the number of turns per conversation ranged from 30-275 turns.  We used a
multilayer perceptron (MLP) with 2 hidden layers of 512 output nodes each to
train the user simulator.  We also introduce regularization with $\ell_2$-norm penalty to constrain model complexity. Because we utilize preset questions by
the interviewer, we use one-hot encoding of questions, denoted $\mathbf{q}_t^i
\in \IR^{d}$, as input for training.
Given the original skip-thought vector
$\mathbf{v}_t^i$, the user simulator serves as a function which maps $f:
\mathbf{q}_t^i \mapsto \mathbf{v}_t^i$. The output of the MLP is the skip-
thought embedding representation of the utterance, denoted $f(\mathbf{q}_t^i;
\vw_i) \in \IR^{c}$. Here, $d$ denotes the size of our question dictionary,
$c$ denotes the dimension of skip-thought embeddings, $\vw_i$ parameterizes
the MLP model for the given user, $i \in N$ denotes the user index and $t \in
T$ denotes the turn number. The loss function of the MLP is given by the mean-
squared error (MSE) between the MLP output and original skip-thought vector
$\mathbf{v}_t^i \in \IR^{c}$:
\begin{align*}
	L(\vw_i) = \frac{1}{2}\ \sum\nolimits_{t=1}^T \left[f(q_t^i;\vw_i) - \mathbf{v}_t^i \right]^2 + \frac{\lambda}{2} ||\vw_i||_2, \hspace{1mm} \forall\ i=1,...,N
\end{align*}
In the case where questions are not preset, 
more state-of-art methods such as end-to-end recurrent neural network systems can be deployed to train the user simulator 
instead ~\cite{wen2016network,li2016user}. To evaluate the performance of our user simulator, 
we computed the mean squared error on the outputs of the simulator and the original thought vector representation 
of the user response for each turn. 
\subsection{Reinforcement Learning Components}
Again, let $c$ denote the size of skip-thought embeddings and $d$ denote the size of question dictionary. We formulate the dialogue and task managers portions of the dialogue system into a standard RL setting
where an agent interacts with environment $\cE$ 
over a finite number of steps. At time step $t$,
the agent receives a state $\vs_t$ and samples an action (asks a question)
$\va_t$ based on its current policy $\pi$. 

The environment transitions to the next state $\vs_{t+1}$ and the agent receives a
scalar reward $r_{t+1}$. In this setting, the RL-agent tries to learn an
optimal policy $\pi^*$ over all possible states, including ones that are
unseen by the agent during training. To do this, the agent has to learn an
approximate action-value function, which maps state-action pairs to expected
future rewards~\cite{sutton1998reinforcement}. 
Formally, the action-value function is defined as follows:
\begin{align*}
	&Q^\pi(\vs,a) = \E_{\pi}\left[\sum\nolimits^{T}_{t=1} \gamma^t r_t|\vs, a\right],
\end{align*}
where $\gamma \in [0, 1]$ is a discount factor and $T$ is the max \# of
turns. 

\noindent \textbf{Environment $\cE$: }  The environment in this case consists
of the dialogue manager (DM), user simulator and MCI classifier. 
DM is composed of the reward and state generating functions.
In previous works, a task manager, composed of a
database and a query manager ~\cite{chen2017survey,wen2016network}, is used by the DM to generate 
observations in retrieval tasks. In our case, however, the the user simulator and MCI classifier is equivalent 
to the task manager and is used by the DM to generate observations. 
Here, the DM uses the MCI classifier to (1) predict probabilities for both the MCI and the NL classes based on current moving-average of skip-thought vectors at each turn, and (2) predict the label of the
current user at the end of the episode for reward calculation. The result of (1) is also used by the agent as part of its internal state-representation. The result of (2) is used by the DM for credit assignment for the
generated conversational episode. The MCI classifier is trained separately on
the training set corpus before the dialogue system phase.

\noindent \textbf{Action $\va_t \in \RR^{d}$: } The RL-agent chooses its actions from a set of 
discrete actions consisting of $107$ predefined questions, where each question is represented by a 
one hot vector in $\RR^{d}$. It is worth noting that we use $\va_t$ and $\vq_t$ 
to differentiate the action taken by our RL-agent and the questions asked during the actual interviews, respectively.

\noindent \textbf{State $\vs_t \in \RR^{C}$:}  
The state representation by the RL-agent is used to approximate the action-value function. 
There are five main components of the state representation vector: 
\begin{itemize}
 \item Skip-thought vector of utterance at current turn: $f(a_{t-1};\vw_i)$, which is the output vector from user simulator $f$ given action $a_{t-1}$ at turn t. 
 \item Moving average of skip-thought vectors across all utterances in current episode: $\bar{f}_{t} = \frac{1}{t} \sum_{k=1}^{t-1} f(a_{k}; \vw_i)$
 \item First hidden layer weights of user-simulator: $\vw_{i}[:,1]$
 \item Predicted probability of current user for MCI and NL classes by classifier
 \item Number of turns above threshold: $\tau$. 
\end{itemize}

The total dimension of the state vector is $C = 2c + |\vw_i[:,1]| + 3 = 10115$.
At each turn, the DM queries the
MCI classifier to output a probability vector composed of 
$P(y_i =0|\bar{f}_t)$ and $P(y_i = 1|\bar{f}_{t})$, 
where $y=0$ denotes NL and $y=1$ denotes MCI. This 2-dimensional vector keeps track
of the classifier\textquotesingle s confidence-level for MCI prediction based
on the current moving-average of skip-thought vectors generated from $1, 2,
..., t$ turns. Keeping track of classifier confidence incentivizes the RL-agent 
to terminate the conversation as soon as it reaches a threshold level of
confidence for the prediction task.


\noindent \textbf{Reward $r \in \RR$:} 
Since we want to minimize the number of dialogue turns, we designed the
environment to output a negative reward (-10) at every time step unless it
reach a terminal state (e.g. when agent says ``goodbye''). At the terminal
state, the reward depends on the classification using the averaged skip-thought 
vector collected from this episodes. If the existing classifier is
able to make the correct prediction, the agent receives a positive reward (1000),
otherwise it receives a moderately negative reward (-500). We also set the
maximum length of episodes $T=35$.  Additionally, we added a linearly
increasing penalty for each passing turn where the classifier predicts with
$\geq0.65$ probability for either class (MCI/NL). We denote this penalty
threshold as the number of turns above confidence threshold ($\tau$).
Formally, the reward function is defined as:
\begin{align}
  r =\left\{
        \begin{array}{ll}
          -10 - 10 \tau,   \hspace{3mm} \text{for non-terminal state}, \\
          -500, \hspace{10mm} \text{terminal state with misclassification}, \\
          +1000, \hspace{8mm} \text{terminal state with correct prediction}.
        \end{array}
      \right.
\label{eq:reward}
\end{align}

\noindent \textbf{State transitions:} 
The state transition function has two parts:
\begin{itemize}[leftmargin=0.1in]
	\item \textit{Within User. } The state transition rule between turns is characterized by:
	\begin{align*}
		P^\pi_{s,s'}&= \sum\nolimits_{a \in A} P(s_{t+1} = s' | s_t = s, a_t = a, \pi) \\
		&= \sum\nolimits_{a \in A} \pi(a|s) P^a_{s,s'}
	\end{align*}
	Given a policy $\pi$, the probability of the environment transitioning to state $s'$ at $s_{t+1}$ depends only on current state $s_t$. Internally, the DM utilizes the user simulator to generate skip-thought $f(a_t; \vw_i)$ from $a_t$.
	\item \textit{Between Users. } In addition to state transitions within episodes, the state-generating function changes between users, leading to different transition probabilities between similar states among different users. To capture this, we apply two changes when training the RL-agent on multiple users: (1) the first hidden layer weights $\vw_i[:,1]$ of each user are incorporated in the state representation vector so that the RL-agent can distinguish between dissimilar users. When used this way, the user simulator provides a means for the RL-agent to learn similar policies for similar users and dissimilar policies for dissimilar users. (2) During training, both the user simulator and classifier of the training environment is reset between users by re-initializing the user simulator weights $\vw_i$ to correspond to the new user. 
\end{itemize}

\noindent \textbf{Deep Q-Networks (DQN).}  
In this work, the action value
function needs to estimate expected reward based on the high-dimensional
state representations as described in previous section. In order to
approximate the action value given different users and the complicated
internal state changing during the conversation,  we learn a deep $Q$-network
parameterized by $\theta_v$ to tackle this challenging problem. The
learning procedure can be conducted by optimizing the loss function as follows:
\begin{align}
L(\theta_v) &= \EE_{\vs_t, \va_t, r_t, \vs_{t+1}'} [(y_t - Q(\vs_{t}, \va_t; \theta_v))],  
\label{eq:dqn}
\end{align}
with   
\begin{align} 
y_t = r_t  + \gamma \max\nolimits_{\va_{t+1}'} Q(\vs_{t+1}', \va_{t+1}'; \theta_v'),
\label{eq:ytarget}
\end{align} 
where $\theta_v'$ denotes the parameters of target $Q$-network. 
In order to learn the estimator under complex situations, 
two key ingredients were proposed in~\cite{mnih2015human}: 
{\it{experience replay}} and fixed {\it{target $Q$-network}}.
During the training, the $Q$-network ($\theta_v$) is updated 
in an online fashion by conducting the gradient descent of \eq{\ref{eq:dqn}}
while the target $Q$-network ($\theta_v'$) is fixed to compute
the target values as in \eq{\ref{eq:ytarget}} and only updated after
a certain number of iterations, which is essential to the 
convergence of $Q$-network in this work. 
We also observe when the {\it{experience replay}} samples minibatch
from previous experiences to update the $Q$-network, the training performance stabilizes more consistently. 



\noindent \textbf{ Policy-masking. } One challenge in our problem is creating an environment that can 
train the agent to produce responses which best align with the flow of conversations. 
For example, an agent may learn that the question ``\textit{can you elaborate on that?} '' is useful for generating a wide distribution of words from the user, but it would not make sense to include that in the first sentence of a conversation or before relevant topics are introduced. 
To achieve this, we created a policy-modifying function in which \textit{confirmation} and \textit{clarification} type questions are masked from the policy set $\pi$ at
turn $t$ if the action history of the agent from $1, 2, ..., t-1$ does not
include any questions from \textit{social}, \textit{activity}, \textit{tech},
\textit{picture-related}, \textit{hobbies}, \textit{occupation},
\textit{travel}, \textit{entertainment} and \textit{family} categories. At
each turn, we keep track of an action history vector $\pi_{t} \in \IR^{d}$ and
construct a policy-masking vector $\varphi_t \in \IR^{d}$ to be applied
element-wise over the agent\textquotesingle s Q-value output. Specifically:
\begin{align}
  &\varphi_t^j =\left\{
        \begin{array}{ll}
          0,   \hspace{3mm} \text{if action $j$ masked}, \\
          1 ,  \hspace{3mm} \text{otherwise}.
        \end{array}
      \right.
\label{eq:mask}
\\
& Q'(\vs_t) = \varphi_t \odot Q(\vs_t). \nonumber
\end{align}
where the $\varphi_t^j$ denotes the $j$-th element in policy-masking vector $\varphi_t$.
And $Q(\vs_t) \in R^d$ represents the action values of all 107 available actions
given current state $\vs_t$. Then the $Q'(\vs_t)$ is valid action values vector 
after the policy masking. 
To achieve effective masking, we assure the elements of 
$Q(\vs_t)$ is positive by using ReLU~\cite{he2016identity} as the activation function for the output
layer of Q-network and a step of pre-training on Q-network as described in following section.


\subsection{Training the RL-Agent} 
We outline below the training procedure for our RL-agent. To expedite the
learning process, we first train the RL-agent over the original corpus from
the training set. For each user, we perform an initial pass through the entire
corpus using the existing action history $q_i^1, q_i^2, ... q_i^T$ to generate
episodes $s_1, a_1, r_1, ... a_t, r_t$. We use these corpus-generated episodes
to train the Q-estimator network. This initialization procedure is motivated
by previous studies which have cited the effectiveness of pre-training with
successful episodes so that the RL-agent can discover large terminal reward
signals in games with delayed rewards ~\cite{anderson2015faster}.

\begin{algorithm}[t!]
\caption{RL-Training Protocol}
\label{alg:env_steps}\small
\begin{algorithmic}
\STATE Initialize replay memory $\mathcal{D}$
\STATE Initialize Task Manager with classifier
\STATE Pre-train action-value function $Q$ 
\FOR {$i = 1,..., N$}
\STATE Initialize Environment $\cE$ with User Simulator $f_i$
\STATE Initialize $\cE$ with true label for user $i$
\FOR {$episode = 1, ..., M$}
\STATE Reset $\cE$ Get the initial state $\vs_1$. 
\FOR {$t = 1, ..., T$}
\STATE Obtain policy mask $\varphi_t$ as~\eq{\ref{eq:mask}}.
\STATE With probability $\epsilon$ select a random action $a_t$ \\ 
otherwise select $a_t = \text{max}_a\  \varphi_t \odot  Q(s_t, a; \theta_v)$
\STATE Execute action $a_t$ in $\cE$ observe reward $r_t$ and state $\vs_{t+1}$
\STATE Store transition $(\vs_t, \va_t, r_t, \vs_{j+1})$ in $\mathcal{D}$
\STATE Sample random minibatch of $(\vs_j, \va_j, r_j, \vs_{j+1})$ from $\mathcal{D}$
	\IF {terminal $\vs_{j+1}$}
		\STATE $y_j = r_j$
	\ELSIF {non-terminal $\vs_{t+1}$}
		\STATE $y_j =  r_j + \gamma \text{max}_a' \ Q(\vs_{j+1}, a'; \theta_v')$
		\ENDIF
\STATE Perform a gradient descent step on $(y_j - Q(\vs_j; a_j, \theta_v))^2$
\ENDFOR
\ENDFOR
\ENDFOR
\end{algorithmic}
\end{algorithm}

During training, we stabilize the target Q-network $\theta_v'$ for minibatch generation and transfer weights from learning Q-network $\theta_v$ every 50 conversational episodes. 
During testing, we use the RL-agent to generate new actions for each test set user $a_1^i, a_2^i, ... a_t^i$. New episodes are then generated by each user simulator from each new action set $\pi_i$ for prediction. These simulated episodes often differ from the original corpus in both the questions asked by the agent as well as in the skip-thought responses by the user.

%% file: sec_pipline.tex

Our proposed framework contains three key learning modules:
 the \emph{user simulator}, the \emph{MCI classifier} and the \emph{RL-agent}. The proposed pipeline is illustrated in Figure~\ref{fig:fig4}. First, the user
simulator is trained by unsupervised learning, which simulates the distributed
representation of user responses given feasible question inputs. Next, the
MCI classifier predicts the patient label based on the averaged
distributed representation of its corpus responses. The above two components
and dialogue manager comprise the training environment for the RL-agent.
The dialogue manager utilizes the user simulator and MCI classifier to handle
the state transitions and also computes of the reward based on the ground-truth 
labels from the training set and MCI classifier prediction.  After training
in this environment, the RL-agent is able to deliver the optimal sequences of
questions for training-set users at various stages of
conversations. During testing, the RL-agent produces query inputs to the test-set user simulators, which represent the unseen users. Using these new queries, the user simulators generate the corresponding distributed representation of test-set user responses for MCI prediction. 
In the following subsections, we will present each component of the 
pipeline in detail and demonstrate the effectiveness of the RL framework in improving prediction accuracy while reducing conversational turns.

%% file: sec_exp.tex
\raggedbottom
Evaluation of dialogue systems differ widely depending on the task. Previous works typically involve using metrics such as perplexity and averaged reward per response to measure the quality of the natural language generation (NLG) phase of the dialogue system~\cite{chen2017survey,wen2016network,schatzmann2006survey}. However, because the utility of our framework  comes from the quality of \textit{questions} that the chat-bot generates for the \emph{off-conversational task}, we propose a framework of evaluation which emphasizes the agent's off-conversation performance. We gauge utility of the dialogue system by its ability to improve (1) prediction accuracy against baseline techniques and (2) the number of turns needed to make accurate prediction. 

\noindent\textbf{Data.}
Data used for this study was obtained from a randomized controlled behavioral clinical trial to ascertain the effect of unstructured conversation on cognitive functions. Details of the study protocol was explained in~\cite{dodge2015web}. In this clinical study, conversational data was collected in Q$\&$A format for each participant during web-cam interviews with trained interviewers. Each participant was interviewed multiple times over the course of 4-6 weeks, and dialogue responses were transcribed for each interview session~\cite{asgari2017predicting}.  On average, there are 2.81 conversational episodes per participant, and each conversation lasted between 30-45 minutes ~\cite{asgari2017predicting,dodge2015web}. MCI labels were generated using clinical assessment of each participant\textquotesingle s cognitive status by medical professionals ~\cite{asgari2017predicting,dodge2015web}. 

\noindent\textbf{Baselines vs. RL Performance.}
We first compare the performance of several baseline classifiers for the MCI prediction task. 
For our specific dataset, ~\cite{asgari2017predicting} had previously achieved benchmark performance of 
72.5$\%$ AUC score on 5-fold validation while using linear SVM with $\ell_1$-norm penalty 
and feature engineering by Linguistic Inquiry and Word Count (LIWC) dictionary ~\cite{asgari2017predicting}. LIWC embeds each word into a 69-dimensional word vector space with each dimension representing a latent feature of the English language ~\cite{asgari2017predicting}. Since 2013, various contextual representations of words and sentences have been proposed, many of which have outperformed classical rule-based contexual embedding techniques ~\cite{mikolov2013distributed,kiros2015skip}. Distributed representation such as \textit{Word2Vec} allows for more flexible and corpus-dependent latent features to be created for individual words~\cite{mikolov2013distributed}. More recently, Skip-thought vectors ~\cite{kiros2015skip} have risen to popularity due to the ability to embed entire sentences into "thought vectors" that capture contextual meaning and syntactic information from neighboring sentences. For this reason, we compare various word and phrase embedding techniques to establish new baseline performances for our classification task.
\begin{table}[t!]
\begin{threeparttable}
\scriptsize

\caption{Performance of basline vs. RL on MCI prediction on 10 stratified shuffle splits} 

\begin{tabular} { |c|c|c|c|c|c| }
	\hline
	\textbf{Model} & \textbf{Feature} & \textbf{AUC} &\textbf{Sen.} & \textbf{Specificity} &\textbf{F1-Score} \\
	\hline
	LR + $\ell_1$ & RD & $0.529\pm0.132$ & $0.380\pm0.260$ & $0.678\pm0.105$ & $0.361\pm0.207$ \\
	RFC & RD & $0.519\pm0.057$ & $0.080\pm0.098$ & $0.944\pm0.075$ & $0.120\pm0.149$ \\
	SVM + $\ell_1$ & RD & $0.551\pm0.131$ & $0.380\pm0.227$ & $0.722\pm0.102$ & $0.384\pm0.214$\\
	SVM + $\ell_2$ & RD & $0.560\pm0.050$ & $0.320\pm0.256$ & $0.800\pm0.185$ & $0.322\pm0.193$\\
	MLP & RD & $0.640\pm0.193$ & $0.110\pm0.243$ & $0.860\pm0.189$ & $0.162\pm0.146$\\
	\hline
	LR + $\ell_1$ & W2V & $0.638\pm0.091$ & $0.520\pm0.204$ & $0.756\pm0.147$ & $0.517\pm0.127$ \\
	RFC & w2v & $0.564\pm0.110$ & $0.340\pm0.220$ & $0.789\pm0.144$ & $0.374\pm0.189$ \\
	SVM + $\ell_1$ & W2V & $0.651\pm0.103$ & $0.560\pm0.233$ & $0.756\pm0.130$ & $0.541\pm0.147$\\
	SVM + $\ell_2$ & W2V & $0.598\pm0.116$ & $0.440\pm0.233$ & $0.756\pm0.171$ & $0.449\pm0.205$\\
	MLP & W2V & $0.680\pm0.151$ & $0.500\pm0.500$ & $0.511\pm0.490$ & $0.266\pm0.266$\\
	\hline
	LR + $\ell_1$ & LIWC & $0.703\pm0.099$ & $0.540\pm0.237$ & $0.867\pm0.130$ & $0.584\pm0.152$ \\
	RFC & LIWC & $0.641\pm0.135$ & $0.360\pm0.250$ & $0.922\pm0.087$ & $0.445\pm0.273$ \\
	SVM + $\ell_1$ & LIWC & $0.661\pm0.125$ & $0.600\pm0.200$ & $0.722\pm0.200$ & $0.572\pm0.144$\\
	SVM + $\ell_2$ & LIWC & $0.712\pm0.110$ & $0.680\pm0.204$ & $0.744\pm0.180$ & $0.631\pm0.135$\\
	MLP & LIWC & $0.689\pm0.129$ & $0.300\pm0.458$ & $0.767\pm0.396$ & $0.182\pm0.285$\\
	\hline
	LR + $\ell_1$ & SKP & $0.790\pm0.112$ & $0.680\pm0.256$ & $0.900\pm0.116$ & $0.707\pm0.183$ \\
	RFC & SKP & $0.608\pm0.104$ & $0.260\pm0.220$ & $0.956\pm0.054$ & $0.343\pm0.259$ \\
	SVM + $\ell_1$ & SKP & $0.783\pm0.123$ & $0.700\pm0.241$ & $0.867\pm0.171$ & $0.711\pm0.190$\\
	SVM + $\ell_2$ & SKP & \textbf{0.797}$\pm\textbf{0.122}$ & \textbf{0.660}$\pm\textbf{0.269}$ & \textbf{0.933}$\pm\textbf{0.102}$ & \textbf{0.716}$\pm\textbf{0.189}$\\
	MLP & SKP & $0.638\pm0.138$ & $0.600\pm0.490$ & $0.400\pm0.490$ & $0.316\pm0.256$\\
	\hline
	RL(T=1) &SKP & 0.607$\pm$0.109 & 0.380$\pm$0.166 & 0.833$\pm$0.134 & 0.447$\pm$0.172 \\
	RL(T=3) & SKP &0.706$\pm$0.092 & 0.500$\pm$0.205 & 0.911$\pm$0.097 & 0.583$\pm$0.154 \\
	RL(T=5) & SKP &0.707$\pm$0.072 & 0.480$\pm$0.133 & 0.933$\pm$0.102& 0.594$\pm$0.129 \\
	RL(T=10) & SKP &0.772$\pm$0.115 & 0.600$\pm$0.237 & 0.944$\pm$0.102 & 0.683$\pm$0.186 \\
	RL(T=15) &SKP & 0.798$\pm$0.115 & 0.640$\pm$0.265 & 0.956$\pm$0.102 & 0.714$\pm$0.190 \\
	RL(T=20) &SKP & 0.798$\pm$0.121 & 0.640$\pm$0.250 & 0.956$\pm$0.102 & 0.719$\pm$0.190 \\
	RL(T=25) & SKP &0.808$\pm$0.111 & 0.660$\pm$0.254 & 0.956$\pm$0.102 & 0.732$\pm$0.184 \\
	RL(T=30) &SKP & 0.808$\pm$0.119 & 0.660$\pm$0.269 & 0.956$\pm$0.102 & 0.730$\pm$0.190 \\
	RL(T=35) &SKP & \textbf{0.818}$\pm$\textbf{0.102} & \textbf{0.680}$\pm$\textbf{0.204} & \textbf{0.956}$\pm$\textbf{0.102} & \textbf{0.761}$\pm$\textbf{0.140} \\
\hline
\end{tabular}
	\begin{tablenotes}
	\scriptsize
	\item Here, \emph{LR} denotes sparse logistic regression classifier, \emph{RFC} denotes random forest classifier, \emph{SVM} denotes support vector machines, and \emph{MLP} denotes multi-layer perceptron. For feature representation of corpus, \emph{RD} represents raw distribution of word counts. \emph{w2v} denotes averaged 300-dimension \emph{Word2Vec} embeddings across all words appearing in the corpus for each user ~\cite{mikolov2013distributed}. \emph{LIWC} denotes the original rule-based embedding used by ~\cite{asgari2017predicting}. \emph{SKP} denote averaged 4800-dimension Skip-Thought vectors across all turn-based responses for each user ~\cite{kiros2015skip}.
	\end{tablenotes}
	\label{table:1}
	\end{threeparttable}
\end{table}

The first four sections of Table \ref{table:1} show the performance of these baseline classifiers. Using the original LIWC representation, we were able to recover close to the 72.5$\%$ AUC baseline from the original paper using SVM and LR classifiers. When implementing skip-thought embedding, we used pre-trained skip-thought encoders by ~\cite{kiros2015skip} to embed each user response across all conversational turns. The encoder was pre-trained on the \textit{BookCorpus} dataset, which is a large collection of novels pertaining to numerous literary genres. The advantage of pre-training on this dataset is that \textit{BookCorpus} contains an abundant number of turn-based dialogues between various character types. These conversations capture a wide range of conversational response styles, idiosyncrasies and temperaments. As seen in Table \ref{table:1}, the best performing baseline model was the SVM classifier with $\ell_2$ norm, using Skip-Thought embedding as features. For this reason, we choose this classifier for the RL portion of our pipeline. As a baseline reference, we also included performance using raw word count distributions for all models. 

We then evaluate the performance of our RL-agent across 10 stratified shuffle splits. Each split uses 65$\%$ of data for training and 35$\%$ for testing. We compare the performance of RL-Agent when manually restricting the number of questions to 1, 3, 5, 7, 10, 15, 20, 25, 30 and $35$. By restricting the number of turns, we can observe the number of questions needed to recover the original baseline performance using the SVM classifier. 
\begin{figure}[t!]
\centering
\includegraphics[width=0.48\textwidth]{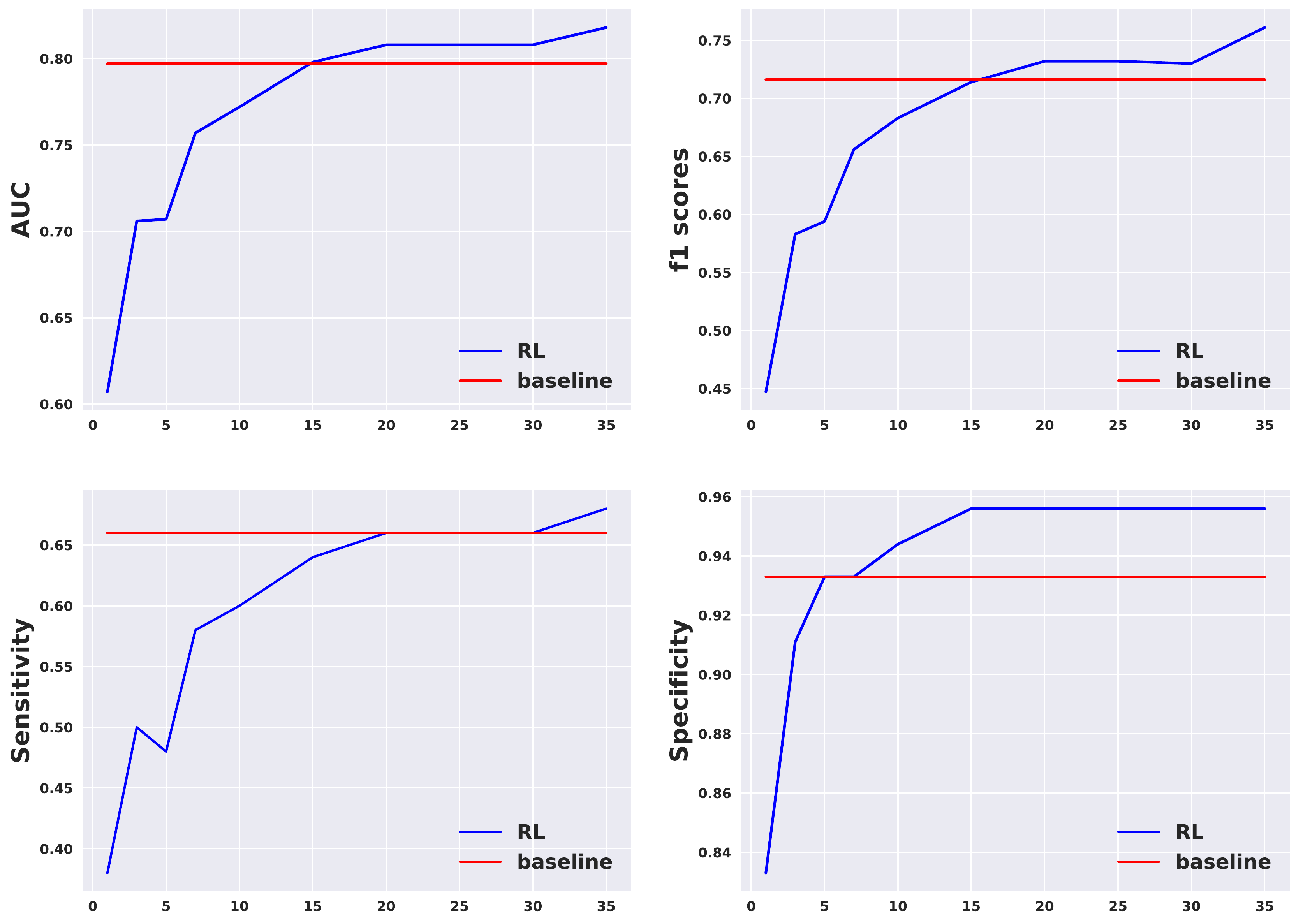}
\vspace{-0.25in}
\caption{RL-Agent vs. Baseline w/ Variation on Turns}
\label{fig:fig7}
\vspace{-0.2in}
\end{figure}

The last section of Table \ref{table:1} illustrates the performance of the RL-agent under various turn constraints. Here, the turns notation RL(T=$t$) denote the number of questions the agent is allowed to ask before a prediction is made from the simulated user responses. It is important to note that turn 0 was set to greetings by default and was not counted toward the conversation. 

We see from constraint conditions that the performance of our RL-agent started to surpass baseline performances starting at 25 questions and was able to achieve comparable performance using only 15 questions. At full conversation length of 35 turns, we were able to achieve 0.818 AUC, an improvement upon current and previous baselines. In comparison, the mean number of conversational turns per user in the original corpus was 105.71. Additionally, since 2.81 conversations were conducted per user, we adjusted the number of turns allowed based on the mean number of turns \textit{per conversation}, which was 37.36 per user. For this reason, we set the upper bound constraint to 35 questions, which is just slightly less than a full conversation with the user. 

Figure \ref{fig:fig7} visualizes this relationship between performance and number of questions asked by the RL-agent. We see that performance improvements with additional questions saturate after 15 questions. This was expected, as the highest-yield questions discovered by the RL-agent were asked first during test conversations.

\noindent\textbf{Evaluation of User Simulators.}
User simulators serve a pivotal role of simulating the user response in the RL training environment ~\cite{schatzmann2006survey,li2016persona}. In previous works, the user simulators are evaluated based on accuracy of generated user query to unseen bot responses ~\cite{li2016persona,schatzmann2006survey}. Metrics such as BLEU and perplexity are used at the NLG phase of dialogue, as the generation of user query is pivotal in retrieval-type training systems. 

In our case, however, the goal of the user simulator is quite different; the RL-agent is responsible for generating queries while the output from the user simulator is actually \textit{an encoded thought-vector} of the user response, which is then used for state representation and downstream prediction purposes. For this reason, we evaluate the performance of the user-simulator not on the decoding portion of the dialogue system, but rather on the performance of the user-simulator in generating accurate thought-vector version of the responses.
\begin{figure}[t!]
\centering
\vspace{-0.2in}
\includegraphics[width=0.25\textwidth]{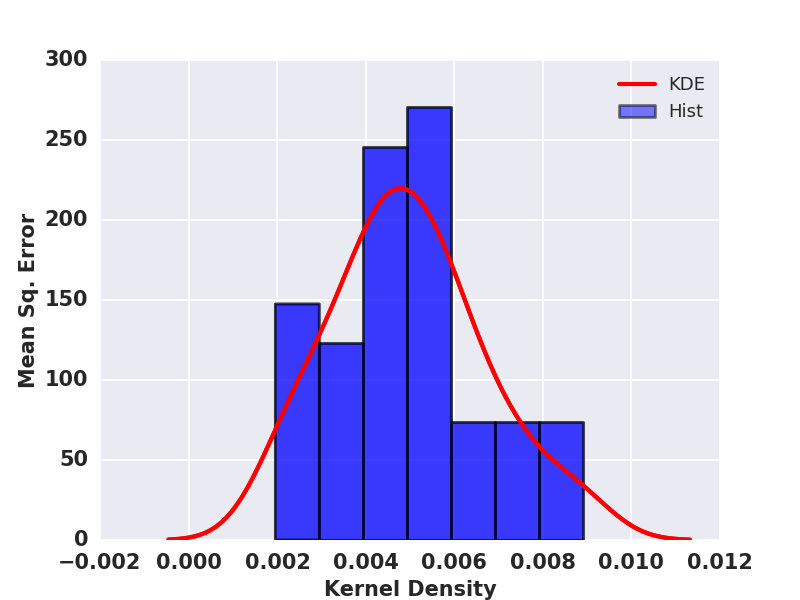}
\vspace{-0.2in}
\caption{Distribution of mean squared error (MSE) across all user simulators.}
\label{fig:fig8}
\vspace{-0.15in}
\end{figure}

We compute mean-squared error (MSE) between the corpus Skip-Thought vector and user simulation prediction at each turn. The resulting MSE scores are averaged across all turns for the conversation. Given that each user has on average 2.81 conversations, we evaluate the performance of the user simulator in a leave-one-out fashion: for each user, the simulator is trained on all conversations except for the last one, which is used for evaluation. Figure \ref{fig:fig8} visualizes the performance of user simulators. The mean MSE is 0.00495$\pm$2.93E-06, averaged across all test set performances.

\noindent\textbf{Top-Performing Policies.}
It is interesting to note that the simulated episodes by our RL-agent were able to provide a performance boost for the prediction task. 
In this section, we look qualitatively at the types of questions at 5, 10, 15, 20 and 35 turns by RL-agent in comparison with the original corpus. We also compare the performance of $\pi^*$@5, @10, @20, @30 and @35 with the performance using the first 5, 10, 20, 30 and 35 responses of the original corpus. Again, we note that responses to greeting and parting queries such as ``\textit{Hi}'' and ``\textit{goodbye}'' are not counted toward prediction. 
\begin{table}[t!]
\Small
\centering
\caption{Prediction @5, 10, 20, 30 and 35 Turns}
\vspace{-0.15in}
\begin{tabular} {|c|c|c|c|c|}
\hline
\textbf{Model} & \textbf{AUC} & \textbf{Sen} &\textbf{Spec} &\textbf{F1-Score} \\ \hline
Corpus@5 & 0.504$\pm$0.070 & 0.120$\pm$0.098 &0.889$\pm$0.099 &0.175$\pm$0.145 \\
Corpus@10 & 0.513$\pm$0.076 & 0.160$\pm$0.174 &0.867$\pm$0.130 &0.193$\pm$0.200 \\
Corpus@20 & 0.614$\pm$0.077 & 0.340$\pm$0.254 &0.889$\pm$0.131 &0.382$\pm$0.223 \\
Corpus@30 & 0.658$\pm$0.121 & 0.360$\pm$0.233 &0.956$\pm$0.056 &0.460$\pm$0.266 \\
Corpus@35 & 0.699$\pm$0.125 & 0.420$\pm$0.244 &0.978$\pm$0.044 &0.539$\pm$0.248 \\
\hline
$\pi^*$@5 & 0.707$\pm$0.072 & 0.480$\pm$0.133 & 0.933$\pm$0.102& 0.594$\pm$0.129 \\
$\pi^*$@10 & 0.772$\pm$0.115 & 0.600$\pm$0.237 & 0.944$\pm$0.102 & 0.683$\pm$0.186 \\
$\pi^*$@20 & 0.798$\pm$0.121 & 0.640$\pm$0.250 & 0.956$\pm$0.102 & 0.719$\pm$0.190 \\
$\pi^*$@30 & 0.808$\pm$0.119 & 0.660$\pm$0.269 & 0.956$\pm$0.102 & 0.730$\pm$0.190 \\
$\pi^*$@35 & \textbf{0.818}$\pm$\textbf{0.102} & \textbf{0.680}$\pm$\textbf{0.204} & \textbf{0.956}$\pm$\textbf{0.102} & \textbf{0.761}$\pm$\textbf{0.140} \\
\hline
\end{tabular}
\label{table:4}
\vspace{-0.2in}
\end{table} 

As shown in Table \ref{table:4}, the optimal policy $\pi^*$ learned by our framework outperformed the original corpus for each turn constraint. For example, when our RL-agent asked only 5 questions to test set users, the classifier was able to achieve 0.707 AUC and 0.594 F1 using the simulated response. In contrast, using the first 5 questions from the original corpus for each test set user produced 0.504 AUC and 0.175 F1. When using the first full-length conversation with 35 turns, the original corpus recovers an AUC score of 0.699, which is far from the performance of $\pi^*$@35. In Table \ref{table:5}, we rank the most frequently appearing questions in $\pi^*$@5, $\pi^*@10$ and $\pi^*@20$.
\begin{table}[t!]
\small
\centering
\caption{Most frequently questions in $\pi^*$@5, 10, 15 and 20}
\vspace{-0.18in}
\begin{tabular}{c|c|c}
\text{Turns} & \text{Question} & {Count} \\ \hline \hline
1-5& when did you start working? & 40 \\
1-5& so how long did you go out for? & 37 \\
1-5& when did you meet your SO? & 28 \\
1-5& <unspecified hobby comment> & 24 \\
1-5& what did you like about <activity>? & 24 \\ 
\hline
6-10& what was <occupation> like for you? & 30 \\
6-10& <unspecified tech comment> & 28 \\
6-10& when did <tech problem> start? & 22 \\
6-10& what do you see in this picture? & 19 \\
6-10& <unspecified picture comment> & 19 \\ \hline
10-15& what is your opinion on <social topic>? & 42 \\
10-15& did you see any shows lately? & 38 \\
10-15& how many people do you think can fit in this? & 33 \\
10-15& what you were doing during this time period? & 30 \\
10-15& what type of <hobby> do you do? & 28 \\
\hline 
15-20& <goodbye> & 27 \\
15-20& where did you meet your so? & 25 \\
15-20& did you enjoy school?& 24 \\
15-20& anyone visit you lately? & 24 \\
15-20& what was the show about?  & 20 \\ 
\end{tabular}
\label{table:5}
\vspace{-0.18in}
\end{table}

\vspace{+0.02in}
\noindent \textbf{$\pi^*$@5. } 
The most effective question in $\pi^*$@5 appears to be ``\textit{when did you start working}''. In the context of our problem, this question seems to generate the most polarizing responses from the cohort. We also see that the RL-agent included a few elaboration questions such as ``\textit{what did you like about <activity>}'' and ``\textit{why did you do that},'' for some users to expand upon previous responses. From the clinical perspective, it is also interesting to note that the RL-agent picks questions such as ``\textit{what did you do yesterday}'' and ``\textit{how long did you go out for},'' which are similar to questions used clinically to assess immediate recall in MCI patients ~\cite{folstein1975mini}. 

\vspace{+0.02in}
\noindent \textbf{$\pi^*$@10. }
As seen in $\pi^*$@5, occupational questions were the most popular topic asked by the RL-agent. This is also the case with $\pi^*$@10, where the RL-agent follows up the previous query with an elaboration question regarding past occupational experiences. It is interesting to note that the RL-agent transitions to picture-related questions, which are often used by the clinical interviewers to facilitate creative responses by participants~\cite{asgari2017predicting}. 

We also observe the RL-agent asking questions such as ``\textit{<unspecified tech comment>}'' and ``\textit{when did <tech problem> start}''. These were frequently asked questions during the course of the original dialogue, as technical difficulties were often encountered with connection and webcam issues during the interviews ~\cite{asgari2017predicting}. Unfortunately, the responses vary greatly and may at times generate verbose responses from participants. The RL-agent did not seem to be able to recognize this caveat during training. 

\vspace{+0.02in}
\noindent \textbf{$\pi^*$@20. } As we approach questions 11 through 20, we arrive at mid- to late- dialogue for most conversations. 
Overall, we observe more widespread topics during this portion of conversation. The most polarizing question asked at this stage was ``\textit{what is your opinion on <social topic>?}'' 
Here, we used delexicalised slots~\cite{liu2017end} <social topic> to reduce model complexity, but the slots may be substituted with a wide range of social topics from political trends to recent news. 

Additionally, we observe that the RL-agent learns to say ``\textit{goodbye}'' to terminate the conversation early in numerous cases. As mentioned previously, we designed the state function to include the predicted probability [0.0-1.0] of MCI by the classifier at each time-step. The environment penalizes the agent for additional turns in which the prediction probability exceeds 0.65 for either class. By opting to terminate the episode, the RL-agent learns to avoid dragging on the dialogue unnecessarily in cases where it is confident in the prediction. 

One notable question in $\pi^*$@20 is ``\textit{how many people do you think can fit in this?}'' This is actually a picture-specific question related to one of the more provocative pictures. In fact, we confirmed from the original corpus that it generated more follow-up response from users when compared to other picture-related questions such as ``\textit{when do you think this picture was taken?}'' and ``\textit{interesting, what makes you say that?}''. By ranking this question highly, the RL-agent indirectly prioritizes this picture over others in generating user responses. This exemplifies how the ranking of questions by $\pi^*$ may be used to direct future data collection process.
 
\vspace{+0.02in}
\noindent \textbf{$\pi^*$@35. } When approaching the end of conversations, we notice that the questions asked by the agent were more spread-out among the remaining choices. For this reason, we rank only the top 10 questions during the final 15 turns of simulated conversations.

{\small
\begin{tabular}{c|c|c}
\text{Rank} & \text{Question} & {Count} \\ \hline \hline
1& what is your opinion on using <new tech>? & 112 \\
2& did you do anything else?  & 106 \\
3& so how long did you go out for? & 98 \\
4& what you were doing during this time period? & 95 \\
5& when do you think this picture was taken? & 95 \\
6& <goodbye> & 94 \\
7& anything new with you lately? & 91 \\
8& what did you like about it? & 85 \\
9& <unspecified picture comment> & 76 \\
10& how often do you <do activity>?  & 72 \\ 
\end{tabular}}
\\ \\ 
In this latter portion of $\pi^*$, we note that the RL-agent utilized more elaboration questions such as ``\textit{what do you like about it}'' and ``\textit{how often do you <do activity>}''. We also see that technology-related questions such as ``\textit{what is your opinion on using <new tech>}'' are included more often when compared to topics such as occupation or social items. This indicates that tech-related questions may not be as high-yield in distinguishing MCI responses, as these questions are prioritized later during conversation by the RL-agent.
